# Rao-Blackwellised Particle Filtering for Dynamic Bayesian Networks

**Arnaud Doucet**[‡]    **Nando de Freitas**[†]    **Kevin Murphy**[†]    **Stuart Russell**[†]

[‡] Engineering Dept.
Cambridge University
ad2@eng.cam.ac.uk

[†] Computer Science Dept.
UC Berkeley
{jfgf,murphyk,russell}@cs.berkeley.edu

## Abstract

Particle filters (PFs) are powerful sampling-based inference/learning algorithms for dynamic Bayesian networks (DBNs). They allow us to treat, in a principled way, any type of probability distribution, nonlinearity and non-stationarity. They have appeared in several fields under such names as "condensation", "sequential Monte Carlo" and "survival of the fittest". In this paper, we show how we can exploit the structure of the DBN to increase the efficiency of particle filtering, using a technique known as Rao-Blackwellisation. Essentially, this samples some of the variables, and marginalizes out the rest exactly, using the Kalman filter, HMM filter, junction tree algorithm, or any other finite dimensional optimal filter. We show that Rao-Blackwellised particle filters (RBPFs) lead to more accurate estimates than standard PFs. We demonstrate RBPFs on two problems, namely non-stationary online regression with radial basis function networks and robot localization and map building. We also discuss other potential application areas and provide references to some finite dimensional optimal filters.

## 1 INTRODUCTION

State estimation (online inference) in state-space models is widely used in a variety of computer science and engineering applications. However, the two most famous algorithms for this problem, the Kalman filter and the HMM filter, are only applicable to linear-Gaussian models and models with finite state spaces, respectively. Even when the state space is finite, it can be so large that the HMM or junction tree algorithms become too computationally expensive. This is typically the case for large discrete dynamic Bayesian networks (DBNs) (Dean and Kanazawa 1989): inference requires at each time space and time that is exponential in the number of hidden nodes.

To handle these problems, sequential Monte Carlo methods, also known as particle filters (PFs), have been introduced (Handschin and Mayne 1969, Akashi and Kumamoto 1977). In the mid 1990s, several PF algorithms were proposed independently under the names of Monte Carlo filters (Kitagawa 1996), sequential importance sampling (SIS) with resampling (SIR) (Doucet 1998), bootstrap filters (Gordon, Salmond and Smith 1993), condensation trackers (Isard and Blake 1996), dynamic mixture models (West 1993), survival of the fittest (Kanazawa, Koller and Russell 1995), etc. One of the major innovations during the 1990s was the inclusion of a resampling step to avoid degeneracy problems inherent to the earlier algorithms (Gordon et al. 1993). In the late nineties, several statistical improvements for PFs were proposed, and some important theoretical properties were established. In addition, these algorithms were applied and tested in many domains: see (Doucet, de Freitas and Gordon 2000) for an up-to-date survey of the field.

One of the major drawbacks of PF is that sampling in high-dimensional spaces can be inefficient. In some cases, however, the model has "tractable substructure", which can be analytically marginalized out, conditional on certain other nodes being imputed, c.f., cutset conditioning in static Bayes nets (Pearl 1988). The analytical marginalization can be carried out using standard algorithms, such as the Kalman filter, the HMM filter, the junction tree algorithm for general DBNs (Cowell, Dawid, Lauritzen and Spiegelhalter 1999), or, any other finite-dimensional optimal filters. The advantage of this strategy is that it can drastically reduce the size of the space over which we need to sample.

Marginalizing out some of the variables is an example of the technique called *Rao-Blackwellisation*, because it is related to the Rao-Blackwell formula: see (Casella and Robert 1996) for a general discussion. Rao-Blackwellised particle filters (RBPF) have been applied in specific contexts such as mixtures of Gaussians (Akashi and Kumamoto 1977, Doucet 1998, Doucet, Godsill and Andrieu 2000), fixed parameter estimation (Kong, Liu and Wong 1994), HMMs (Doucet 1998, Doucet, Godsill and Andrieu 2000) and Dirichlet process models (MacEachern, Clyde and Liu 1999). In this paper, we develop the general theory of RBPFs, and apply it to several novel types of DBNs. We omit the proofs of the theorems for lack of space: please refer to the technical report (Doucet, Gordon and Krishnamurthy 1999).

## 2 PROBLEM FORMULATION

Let us consider the following general state space model/DBN with hidden variables $z_t$ and observed variables $y_t$. We assume that $z_t$ is a Markov process of initial distribution $p(z_0)$ and transition equation $p(z_t|z_{t-1})$. The observations $y_{1:t} \triangleq \{y_1, y_2, \ldots, y_t\}$ are assumed to be conditionally independent given the process $z_t$ of marginal distribution $p(y_t|z_t)$. Given these observations, the inference of any subset or property of the states $z_{0:t} \triangleq \{z_0, z_1, \ldots, z_t\}$ relies on the joint posterior distribution $p(z_{0:t}|y_{1:t})$. Our objective is, therefore, to estimate this distribution, or some of its characteristics such as the filtering density $p(z_t|y_{1:t})$ or the minimum mean square error (MMSE) estimate $\mathbb{E}[z_t|y_{1:t}]$. The posterior satisfies the following recursion

$$p(z_{0:t}|y_{1:t}) = p(z_{0:t-1}|y_{1:t-1}) \frac{p(y_t|z_t) p(z_t|z_{t-1})}{p(y_t|y_{1:t-1})} \quad (1)$$

If one attempts to solve this problem analytically, one obtains integrals that are not tractable. One, therefore, has to resort to some form of numerical approximation scheme. In this paper, we focus on sampling-based methods. Advantages and disadvantages of other approaches are discussed at length in (de Freitas 1999).

The above description assumes that there is no structure within the hidden variables. But suppose we can divide the hidden variables $z_t$ into two groups, $r_t$ and $x_t$, such that $p(z_t|z_{t-1}) = p(x_t|r_{t-1:t}, x_{t-1})p(r_t|r_{t-1})$ and, conditional on $r_{0:t}$, the conditional posterior distribution $p(x_{0:t}|y_{1:t}, r_{0:t})$ is analytically tractable.[1] Then we can easily marginalize out $x_{0:t}$ from the posterior, and only need to focus on estimating $p(r_{0:t}|y_{1:t})$, which lies in a space of reduced dimension. Formally, we are making use of the following decomposition of the posterior, which follows from the chain rule

$$p(r_{0:t}, x_{0:t}|y_{1:t}) = p(x_{0:t}|y_{1:t}, r_{0:t}) p(r_{0:t}|y_{1:t})$$

The marginal posterior distribution $p(r_{0:t}|y_{1:t})$ satisfies

the alternative recursion

$$p(r_{0:t}|y_{1:t}) = \frac{p(y_t|y_{1:t-1}, r_{0:t}) p(r_t|r_{t-1}) p(r_{0:t-1}|y_{1:t-1})}{p(y_t|y_{1:t-1})} \quad (2)$$

If eq. (1) does not admit a closed-form expression, then eq. (2) does not admit one either and sampling-based methods are also required. Since the dimension of $p(r_{0:t}|y_{1:t})$ is smaller than the one of $p(r_{0:t}, x_{0:t}|y_{1:t})$, we should expect to obtain better results.

In the following section, we review the importance sampling (IS) method, which is the core of PF, and quantify the improvements one can expect by marginalizing out $x_{0:t}$, i.e. using the so-called Rao-Blackwellised estimate. Subsequently, in Section 4, we describe a general RBPF algorithm and detail the implementation issues.

## 3 IMPORTANCE SAMPLING AND RAO-BLACKWELLISATION

If we were able to sample $N$ i.i.d. random samples (particles), $\left\{\left(r_{0:t}^{(i)}, x_{0:t}^{(i)}\right); i = 1, \ldots, N\right\}$, according to $p(r_{0:t}, x_{0:t}|y_{1:t})$, then an empirical estimate of this distribution would be given by

$$\overline{P_N}(r_{0:t}, x_{0:t}|y_{1:t}) = \frac{1}{N} \sum_{i=1}^{N} \delta_{\left(r_{0:t}^{(i)}, x_{0:t}^{(i)}\right)} (dr_{0:t} dx_{0:t})$$

where $\delta_{\left(r_{0:t}^{(i)}, x_{0:t}^{(i)}\right)}(dr_{0:t} dx_{0:t})$ denotes the Dirac delta function located at $\left(r_{0:t}^{(i)}, x_{0:t}^{(i)}\right)$. As a corollary, an estimate of the filtering distribution $p(r_t, x_t|y_{1:t})$ is $\overline{P_N}(r_t, x_t|y_{1:t}) = \frac{1}{N} \sum_{i=1}^{N} \delta_{\left(r_t^{(i)}, x_t^{(i)}\right)}(dr_t dx_t)$. Hence one can easily estimate the expected value of any function $f_t$ of the hidden variables w.r.t. this distribution, $I(f_t)$, using

$$\overline{I_N}(f_t) = \int f_t(r_{0:t}, x_{0:t}) \overline{P_N}(r_{0:t}, x_{0:t}|y_{1:t}) dr_{0:t} dx_{0:t}$$

$$= \frac{1}{N} \sum_{i=1}^{N} f_t\left(r_{0:t}^{(i)}, x_{0:t}^{(i)}\right)$$

This estimate is unbiased and, from the strong law of large numbers (SLLN), $\overline{I_N}(f_t)$ converges almost surely (a.s.) towards $I(f_t)$ as $N \to +\infty$. If $\sigma_{f_t}^2 \triangleq \text{var}_{p(r_{0:t}, x_{0:t}|y_{1:t})}[f_t(r_{0:t}, x_{0:t})] < +\infty$, then a central limit theorem (CLT) holds

$$\sqrt{N}\left[\overline{I_N}(f_t) - I(f_t)\right] \underset{N \to \infty}{\Longrightarrow} \mathcal{N}\left(0, \sigma_{f_t}^2\right)$$

where " $\Rightarrow$ " denotes convergence in distribution. Typically, it is impossible to sample efficiently from the "target" posterior distribution $p(r_{0:t}, x_{0:t}|y_{1:t})$ at any time $t$. So we focus on alternative methods.

---

[1] The problem of how to automatically identify which variables should be sampled, and which can be handled analytically, is one we are currently working on. We anticipate that algorithms similar to cutset conditioning (Becker, Bar-Yehuda and Geiger 1999) might prove useful.

One way to estimate $p(\mathbf{r}_{0:t}, \mathbf{x}_{0:t}| \mathbf{y}_{1:t})$ and $I(\mathbf{f}_t)$ consists of using the well-known importance sampling method (Bernardo and Smith 1994). This method is based on the following observation. Let us introduce an arbitrary importance distribution $q(\mathbf{r}_{0:t}, \mathbf{x}_{0:t}| \mathbf{y}_{1:t})$, from which it is easy to get samples, and such that $p(\mathbf{r}_{0:t}, \mathbf{x}_{0:t}| \mathbf{y}_{1:t}) > 0$ implies $q(\mathbf{r}_{0:t}, \mathbf{x}_{0:t}| \mathbf{y}_{1:t}) > 0$. Then

$$I(\mathbf{f}_t) = \frac{\mathbb{E}_{q(\mathbf{r}_{0:t}, \mathbf{x}_{0:t}| \mathbf{y}_{1:t})}(\mathbf{f}_t(\mathbf{r}_{0:t}, \mathbf{x}_{0:t}) w(\mathbf{r}_{0:t}, \mathbf{x}_{0:t}))}{\mathbb{E}_{q(\mathbf{r}_{0:t}, \mathbf{x}_{0:t}| \mathbf{y}_{1:t})}(w(\mathbf{r}_{0:t}, \mathbf{x}_{0:t}))}$$

where the importance weight is equal to

$$w(\mathbf{r}_{0:t}, \mathbf{x}_{0:t}) = \frac{p(\mathbf{r}_{0:t}, \mathbf{x}_{0:t}| \mathbf{y}_{1:t})}{q(\mathbf{r}_{0:t}, \mathbf{x}_{0:t}| \mathbf{y}_{1:t})}$$

Given $N$ i.i.d. samples $\left\{\left(\mathbf{r}_{0:t}^{(i)}, \mathbf{x}_{0:t}^{(i)}\right)\right\}$ distributed according to $q(\mathbf{r}_{0:t}, \mathbf{x}_{0:t}| \mathbf{y}_{1:t})$, a Monte Carlo estimate of $I(\mathbf{f}_t)$ is given by

$$\widehat{I_N^1}(\mathbf{f}_t) = \frac{\widehat{A_N^1}(\mathbf{f}_t)}{\widehat{B_N^1}(\mathbf{f}_t)} = \frac{\sum_{i=1}^N \mathbf{f}_t\left(\mathbf{r}_{0:t}^{(i)}, \mathbf{x}_{0:t}^{(i)}\right) w\left(\mathbf{r}_{0:t}^{(i)}, \mathbf{x}_{0:t}^{(i)}\right)}{\sum_{i=1}^N w\left(\mathbf{r}_{0:t}^{(i)}, \mathbf{x}_{0:t}^{(i)}\right)}$$

$$= \sum_{i=1}^N \widetilde{w}_{0:t}^{(i)} \mathbf{f}_t\left(\mathbf{r}_{0:t}^{(i)}, \mathbf{x}_{0:t}^{(i)}\right)$$

where the normalized importance weights $\widetilde{w}_{1:t}^{(i)}$ are equal to

$$\widetilde{w}_{0:t}^{(i)} = \frac{w\left(\mathbf{r}_{0:t}^{(i)}, \mathbf{x}_{0:t}^{(i)}\right)}{\sum_{j=1}^N w\left(\mathbf{r}_{0:t}^{(j)}, \mathbf{x}_{0:t}^{(j)}\right)}$$

This method is equivalent to the following point mass approximation of $p(\mathbf{r}_{0:t}, \mathbf{x}_{0:t}| \mathbf{y}_{1:t})$

$$\widehat{p_N}(\mathbf{r}_{0:t}, \mathbf{x}_{0:t}| \mathbf{y}_{1:t}) = \sum_{i=1}^N \widetilde{w}_{0:t}^{(i)} \delta_{\left(\mathbf{r}_{0:t}^{(i)}, \mathbf{x}_{0:t}^{(i)}\right)}(d\mathbf{r}_{0:t} d\mathbf{x}_{0:t})$$

For "perfect" simulation, that is $q(\mathbf{r}_{0:t}, \mathbf{x}_{0:t}| \mathbf{y}_{1:t}) = p(\mathbf{r}_{0:t}, \mathbf{x}_{0:t}| \mathbf{y}_{1:t})$, we would have $\widetilde{w}_{0:t}^{(i)} = N^{-1}$ for any $i$. In practice, we will try to select the importance distribution as close as possible to the target distribution in a given sense. For $N$ finite, $\widehat{I_N^1}(\mathbf{f}_t)$ is biased (since it is a ratio of estimates), but according to the SLLN, $\widehat{I_N}(\mathbf{f}_t)$ converges asymptotically a.s. towards $I(\mathbf{f}_t)$. Under additional assumptions, a CLT also holds.

Now consider the case where one can marginalize out $\mathbf{x}_{0:t}$ analytically, then we can propose an alternative estimate for $I(\mathbf{f}_t)$ with a reduced variance. As $p(\mathbf{r}_{0:t}, \mathbf{x}_{0:t}| \mathbf{y}_{1:t}) = p(\mathbf{r}_{0:t}| \mathbf{y}_{1:t}) p(\mathbf{x}_{0:t}| \mathbf{y}_{1:t}, \mathbf{r}_{0:t})$, where $p(\mathbf{x}_{0:t}| \mathbf{y}_{1:t}, \mathbf{r}_{0:t})$ is a distribution that can be computed exactly, then an approximation of $p(\mathbf{r}_{0:t}| \mathbf{y}_{1:t})$ yields straightforwardly an approximation of $p(\mathbf{r}_{0:t}, \mathbf{x}_{0:t}| \mathbf{y}_{1:t})$. Moreover, if $\mathbb{E}_{p(\mathbf{x}_{0:t}| \mathbf{y}_{1:t}, \mathbf{r}_{0:t})}(\mathbf{f}_t(\mathbf{r}_{0:t}, \mathbf{x}_{0:t}))$ can be evaluated in a closed-form expression, then the following alternative importance sampling estimate of $I(\mathbf{f}_t)$ can be used

$$\widehat{I_N^2}(\mathbf{f}_t) = \frac{\widehat{A_N^2}(\mathbf{f}_t)}{\widehat{B_N^2}(\mathbf{f}_t)}$$

$$= \frac{\sum_{i=1}^N \mathbb{E}_{p\left(\mathbf{x}_{0:t}| \mathbf{y}_{1:t}, \mathbf{r}_{0:t}^{(i)}\right)}\left(\mathbf{f}_t\left(\mathbf{r}_{0:t}^{(i)}, \mathbf{x}_{0:t}\right)\right) w\left(\mathbf{r}_{0:t}^{(i)}\right)}{\sum_{i=1}^N w\left(\mathbf{r}_{0:t}^{(i)}\right)}$$

where

$$w(\mathbf{r}_{0:t}) = \frac{p(\mathbf{r}_{0:t}| \mathbf{y}_{1:t})}{q(\mathbf{r}_{0:t}| \mathbf{y}_{1:t})}$$

$$q(\mathbf{r}_{0:t}| \mathbf{y}_{1:t}) = \int q(\mathbf{r}_{0:t}, \mathbf{x}_{0:t}| \mathbf{y}_{1:t}) d\mathbf{x}_{0:t}$$

Intuitively, to reach a given precision, $\widehat{I_N^2}(\mathbf{f}_t)$ will require a reduced number $N$ of samples over $\widehat{I_N^1}(\mathbf{f}_t)$ as we only need to sample from a lower-dimensional distribution. This is proven in the following propositions.

**Proposition 1** *The variances of the importance weights, the numerators and the denominators satisfy for any $N$*

$$\text{var}_{q(\mathbf{r}_{0:t}| \mathbf{y}_{1:t})}(w(\mathbf{r}_{0:t})) \leq \text{var}_{q(\mathbf{r}_{0:t}, \mathbf{x}_{0:t}| \mathbf{y}_{1:t})}(w(\mathbf{r}_{0:t}, \mathbf{x}_{0:t}))$$

$$\text{var}_{q(\mathbf{r}_{0:t}| \mathbf{y}_{1:t})}\left(\widehat{A_N^2}(\mathbf{f}_t)\right) \leq \text{var}_{q(\mathbf{r}_{0:t}, \mathbf{x}_{0:t}| \mathbf{y}_{1:t})}\left(\widehat{A_N^1}(\mathbf{f}_t)\right)$$

$$\text{var}_{q(\mathbf{r}_{0:t}| \mathbf{y}_{1:t})}\left(\widehat{B_N^2}(\mathbf{f}_t)\right) \leq \text{var}_{q(\mathbf{r}_{0:t}, \mathbf{x}_{0:t}| \mathbf{y}_{1:t})}\left(\widehat{B_N^1}(\mathbf{f}_t)\right)$$

A sufficient condition for $\widehat{I_N^1}(\mathbf{f}_t)$ to satisfy a CLT is $\text{var}_{p(\mathbf{r}_{0:t}, \mathbf{x}_{0:t}| \mathbf{y}_{1:t})}\{\mathbf{f}_t(\mathbf{r}_{0:t}, \mathbf{x}_{0:t})\} < +\infty$ and $w(\mathbf{r}_{0:t}, \mathbf{x}_{0:t}) < +\infty$ for any $(\mathbf{r}_{0:t}, \mathbf{x}_{0:t})$ (Bernardo and Smith 1994). This trivially implies that $\widehat{I_N^2}(\mathbf{f}_t)$ also satisfies a CLT. More precisely, we get the following result.

**Proposition 2** *Under the assumptions given above, $\widehat{I_N^1}(\mathbf{f}_t)$ and $\widehat{I_N^2}(\mathbf{f}_t)$ satisfy a CLT*

$$\sqrt{N}\left(\widehat{I_N^1}(\mathbf{f}_t) - I(\mathbf{f}_t)\right) \underset{N \to \infty}{\Longrightarrow} \mathcal{N}(0, \sigma_1^2)$$

$$\sqrt{N}\left(\widehat{I_N^2}(\mathbf{f}_t) - I(\mathbf{f}_t)\right) \underset{N \to \infty}{\Longrightarrow} \mathcal{N}(0, \sigma_2^2)$$

*where $\sigma_1^2 \geq \sigma_2^2$, $\sigma_1^2$ and $\sigma_2^2$ being given by*

$$\sigma_1^2 = \mathbb{E}_{q(\mathbf{r}_{0:t}, \mathbf{x}_{0:t}| \mathbf{y}_{1:t})}\left[((\mathbf{f}_t(\mathbf{r}_{0:t}, \mathbf{x}_{0:t}) - I(\mathbf{f}_t)) w(\mathbf{r}_{0:t}, \mathbf{x}_{0:t}))^2\right]$$

$$\sigma_2^2 = \mathbb{E}_{q(\mathbf{r}_{0:t}| \mathbf{y}_{1:t})}\left[((\mathbb{E}_{p(\mathbf{x}_{0:t}| \mathbf{y}_{1:t}, \mathbf{r}_{0:t})}(\mathbf{f}_t(\mathbf{r}_{0:t}, \mathbf{x}_{0:t})) - I(\mathbf{f}_t)) w_t(\mathbf{r}_{0:t}))^2\right]$$

The Rao-Blackwellised estimate $\widehat{I_N^2}(\mathbf{f}_t)$ is usually computationally more extensive to compute than $\widehat{I_N^1}(\mathbf{f}_t)$ so it is of interest to know when, for a fixed computational complexity, one can expect to achieve variance reduction. One

has

$$\sigma_1^2 - \sigma_2^2 = \mathbb{E}_{q(\mathbf{r}_{0:t}|\mathbf{y}_{1:t})}\left[var_{q(\mathbf{x}_{0:t}|\mathbf{y}_{1:t},\mathbf{r}_{0:t})}\left[(\mathbf{f}_t(\mathbf{r}_{0:t},\mathbf{x}_{0:t}) - I(\mathbf{f}_t))w(\mathbf{r}_{0:t},\mathbf{x}_{0:t})\right]\right]$$

so that, accordingly to the intuition, it will be worth generally performing Rao-Blackwellisation when the average conditional variance of the variable $\mathbf{x}_{0:t}$ is high.

## 4  RAO-BLACKWELLISED PARTICLE FILTERS

Given $N$ particles (samples) $\{\mathbf{r}_{0:t-1}^{(i)}, \mathbf{x}_{0:t-1}^{(i)}\}$ at time $t-1$, approximately distributed according to the distribution $p(\mathbf{r}_{0:t-1}^{(i)}, \mathbf{x}_{0:t-1}^{(i)}|\mathbf{y}_{1:t-1})$, RBPFs allow us to compute $N$ particles $(\mathbf{r}_{0:t}^{(i)}, \mathbf{x}_{0:t}^{(i)})$ approximately distributed according to the posterior $p(\mathbf{r}_{0:t}^{(i)}, \mathbf{x}_{0:t}^{(i)}|\mathbf{y}_{1:t})$, at time $t$. This is accomplished with the algorithm shown below, the details of which will now be explained.

---

**Generic RBPF**

1. Sequential importance sampling step

   - For $i = 1, \ldots, N$, sample:
   $$\left(\widehat{\mathbf{r}}_t^{(i)}\right) \sim q(\mathbf{r}_t|\mathbf{r}_{0:t-1}^{(i)}, \mathbf{y}_{1:t})$$
   and set:
   $$\left(\widehat{\mathbf{r}}_{0:t}^{(i)}\right) \triangleq (\widehat{\mathbf{r}}_t^{(i)}, \mathbf{r}_{0:t-1}^{(i)})$$

   - For $i = 1, \ldots, N$, evaluate the importance weights up to a normalizing constant:
   $$w_t^{(i)} = \frac{p(\widehat{\mathbf{r}}_{0:t}^{(i)}|\mathbf{y}_{1:t})}{q(\widehat{\mathbf{r}}_t^{(i)}|\mathbf{r}_{0:t-1}^{(i)},\mathbf{y}_{1:t})p(\widehat{\mathbf{r}}_{0:t-1}^{(i)}|\mathbf{y}_{1:t-1})}$$

   - For $i = 1, \ldots, N$, normalize the importance weights:
   $$\widetilde{w}_t^{(i)} = w_t^{(i)}\left[\sum_{j=1}^N w_t^{(j)}\right]^{-1}$$

2. Selection step

   - Multiply/ suppress samples $(\widehat{\mathbf{r}}_{0:t}^{(i)})$ with high/low importance weights $\widetilde{w}_t^{(i)}$, respectively, to obtain $N$ random samples $(\widetilde{\mathbf{r}}_{0:t}^{(i)})$ approximately distributed according to $p(\widetilde{\mathbf{r}}_{0:t}^{(i)}|\mathbf{y}_{1:t})$.

3. MCMC step

   - Apply a Markov transition kernel with invariant distribution given by $p(\mathbf{r}_{0:t}^{(i)}|\mathbf{y}_{1:t})$ to obtain $(\mathbf{r}_{0:t}^{(i)})$.

---

### 4.1  IMPLEMENTATION ISSUES

#### 4.1.1  Sequential importance sampling

If we restrict ourselves to importance functions of the following form

$$q(\mathbf{r}_{0:t}|\mathbf{y}_{1:t}) = q(\mathbf{r}_0)\prod_{k=1}^t q(\mathbf{r}_k|\mathbf{y}_{1:k},\mathbf{r}_{1:k-1}) \quad (3)$$

we can obtain recursive formulas to evaluate $w(\mathbf{r}_{0:t}) = w(\mathbf{r}_{0:t-1})w_t$ and thus $\widetilde{w}_{1:t}$. The "incremental weight" $w_t$ is given by

$$w_t \propto \frac{p(\mathbf{y}_t|\mathbf{y}_{1:t-1},\mathbf{r}_{0:t})p(\mathbf{r}_t|\mathbf{r}_{t-1})}{q(\mathbf{r}_t|\mathbf{y}_{1:t},\mathbf{r}_{1:t-1})}$$

$\widetilde{w}_t$ denotes the normalized version of $w_t$, i.e. $\widetilde{w}_t^{(i)} = \left[\sum_{j=1}^N w_t^{(j)}\right]^{-1} w_t^{(i)}$. Hence we can perform importance sampling online.

*Choice of the Importance Distribution*

There are infinitely many possible choices for $q(\mathbf{r}_{0:t}|\mathbf{y}_{1:t})$, the only condition being that its supports must include that of $p(\mathbf{r}_{0:t}|\mathbf{y}_{1:t})$. The simplest choice is to just sample from the prior, $p(\mathbf{r}_t|\mathbf{r}_{t-1})$, in which case the importance weight is equal to the likelihood, $p(\mathbf{y}_t|\mathbf{y}_{1:t-1},\mathbf{r}_{0:t})$. This is the most widely used distribution, since it is simple to compute, but it can be inefficient, since it ignores the most recent evidence, $\mathbf{y}_t$. Intuitively, many of our samples may end up in a region of the space that has low likelihood, and hence receive low weight; these particles are effectively wasted.

We can show that the "optimal" proposal distribution, in the sense of minimizing the variance of the importance weights, takes the most recent evidence into account:

**Proposition 3** *The distribution that minimizes the variance of the importance weights conditional upon $\mathbf{r}_{0:t-1}$ and $\mathbf{y}_{1:t}$ is*

$$p(\mathbf{r}_t|\mathbf{r}_{0:t-1},\mathbf{y}_{1:t}) = \frac{p(\mathbf{y}_t|\mathbf{y}_{1:t-1},\mathbf{r}_{0:t})p(\mathbf{r}_t|\mathbf{r}_{t-1})}{p(\mathbf{y}_t|\mathbf{y}_{1:t-1},\mathbf{r}_{0:t-1})}$$

*and the associated importance weight $w_t$ is*

$$p(\mathbf{y}_t|\mathbf{y}_{1:t-1},\mathbf{r}_{0:t-1}) = \int p(\mathbf{y}_t|\mathbf{y}_{1:t-1},\mathbf{r}_{0:t})p(\mathbf{r}_t|\mathbf{r}_{t-1})d\mathbf{r}_t$$

Unfortunately, computing the optimal importance sampling distribution is often too expensive. Several deterministic approximations to the optimal distribution have been proposed, see for example (de Freitas 1999, Doucet 1998).

*Degeneracy of SIS*

The following proposition shows that, for importance functions of the form (3), the variance of $w(\mathbf{r}_{0:t})$ can only increase (stochastically) over time. The proof of this proposition is an extension of a Kong-Liu-Wong theorem (Kong et al. 1994, p. 285) to the case of an importance function of the form (3).

**Proposition 4** *The unconditional variance (i.e. with the observations $\mathbf{y}_{1:t}$ being interpreted as random variables) of the importance weights $w(\mathbf{r}_{0:t})$ increases over time.*

In practice, the degeneracy caused by the variance increase can be observed by monitoring the importance weights. Typically, what we observe is that, after a few iterations, one of the normalized importance weights tends to 1, while the remaining weights tend to zero.

### 4.1.2 Selection step

To avoid the degeneracy of the sequential importance sampling simulation method, a selection (resampling) stage may be used to eliminate samples with low importance ratios and multiply samples with high importance ratios. A selection scheme associates to each particle $\mathbf{r}_{0:t}^{(i)}$ a number of offsprings, say $N_i \in \mathbb{N}$, such that $\sum_{i=1}^{N} N_i = N$. Several selection schemes have been proposed in the literature. These schemes satisfy $\mathbb{E}(N_i) = N\widetilde{w}_t^{(i)}$, but their performance varies in terms of the variance of the particles, $\text{var}(N_i)$. Recent theoretical results in (Crisan, Del Moral and Lyons 1999) indicate that the restriction $\mathbb{E}(N_i) = N\widetilde{w}_t^{(i)}$ is unnecessary to obtain convergence results (Doucet et al. 1999). Examples of these selection schemes include multinomial sampling (Doucet 1998, Gordon et al. 1993, Pitt and Shephard 1999), residual resampling (Kitagawa 1996, Liu and Chen 1998) and stratified sampling (Kitagawa 1996). Their computational complexity is $\mathcal{O}(N)$.

### 4.1.3 MCMC step

After the selection scheme at time $t$, we obtain $N$ particles distributed marginally approximately according to $p(\mathbf{r}_{0:t}|\mathbf{y}_{1:t})$. As discussed earlier, the discrete nature of the approximation can lead to a skewed importance weights distribution. That is, many particles have no offspring ($N_i = 0$), whereas others have a large number of offspring, the extreme case being $N_i = N$ for a particular value $i$. In this case, there is a severe reduction in the diversity of the samples. A strategy for improving the results involves introducing MCMC steps of invariant distribution $p(\mathbf{r}_{0:t}|\mathbf{y}_{1:t})$ on each particle (Andrieu, de Freitas and Doucet 1999b, Gilks and Berzuini 1998, MacEachern et al. 1999). The basic idea is that, by applying a Markov transition kernel, the total variation of the current distribution with respect to the invariant distribution can only decrease. Note, however, that we do not require this kernel to be ergodic.

### 4.2 CONVERGENCE RESULTS

Let $B(\mathbb{R}^n)$ be the space of bounded, Borel measurable functions on $\mathbb{R}^n$. We denote $\|f\| \triangleq \sup_{x \in \mathbb{R}^n} |f(x)|$. The following theorem is a straightforward consequence of Theorem 1 in (Crisan and Doucet 2000) which is an extension of previous results in (Crisan et al. 1999).

**Theorem 5** *If the importance weights $w_t$ are upper bounded and if one uses one of the selection schemes described previously, then, for all $t \geq 0$, there exists $c_t$ independent of $N$ such that for any $f_t \in B\left((\mathbb{R}^{n_z})^{t+1}\right)$*

$$\mathbb{E}\left[\left(\frac{1}{N}\sum_{i=1}^{N} f_t\left(\mathbf{r}_{0:t}^{(i)}\right) - \int f_t(\mathbf{r}_{0:t}) p(\mathbf{r}_{0:t}|\mathbf{y}_{1:t}) d\mathbf{r}_{0:t}\right)^2\right] \leq c_t \frac{\|f_t\|^2}{N}$$

where the expectation is taken w.r.t. to the randomness introduced by the PF algorithm. This results shows that, under very lose assumptions, convergence of this general particle filtering method is ensured and that the convergence rate of the method is independent of the dimension of the state-space. However, $c_t$ usually increases exponentially with time. If additional assumptions on the dynamic system under study are made (e.g. discrete state spaces), it is possible to get uniform convergence results ($c_t = c$ for any $t$) for the filtering distribution $p(\mathbf{x}_t|\mathbf{y}_{1:t})$. We do not pursue this here.

## 5 EXAMPLES

We now illustrate the theory by briefly describing two applications we have worked on.

### 5.1 ON-LINE REGRESSION AND MODEL SELECTION WITH NEURAL NETWORKS

Consider a function approximation scheme consisting of a mixture of $k$ radial basis functions (RBFs) and a linear regression term. The number of basis functions, $k_t$, their centers, $\boldsymbol{\mu}_t$, the coefficients (weights of the RBF centers plus regression terms), $\boldsymbol{\alpha}_t$, and the variance of the Gaussian noise on the output, $\sigma_t^2$, can all vary with time, so we treat them as latent random variables: see Figure 1. For details, see (Andrieu, de Freitas and Doucet 1999a).

In (Andrieu et al. 1999a), we show that it is possible to simulate $\boldsymbol{\mu}_t$, $k_t$ and $\sigma_t$ with a particle filter and to compute the coefficients $\boldsymbol{\alpha}_t$ analytically using Kalman filters. This is possible because the output of the neural network is linear in $\boldsymbol{\alpha}_t$, and hence the system is a conditionally linear Gaussian state-space model (CLGSSM), that is it is a linear Gaussian state-space model conditional upon the location of the bases and the hyper-parameters. This leads to an efficient RBPF that can be combined with a reversible jump MCMC algorithm (Green 1995) to select the number of basis functions online. For example, we generated some data from a mixture of 2 RBFs for $t = 1, \ldots, 500$, and then from a single RBF for $t = 501, \ldots, 1000$; the method was able to track this change, as shown in Figure 2. Further experiments on real data sets are described in (Andrieu et al. 1999a).

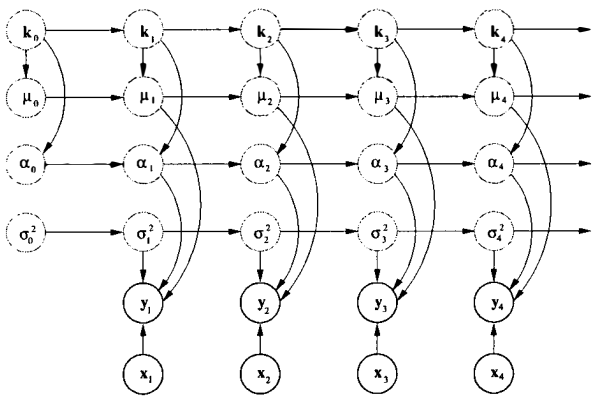

Figure 1: DBN representation of the RBF model. The hyper-parameters have been omitted for clarity.

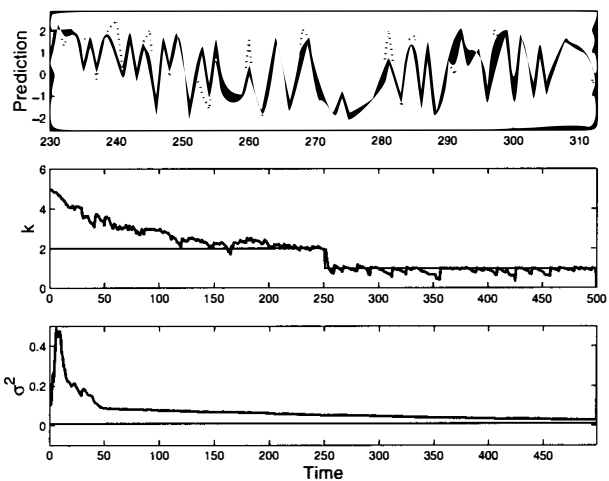

Figure 2: The top plot shows the one-step-ahead output predictions [—] and the true outputs [···] for the RBF model. The middle and bottom plots show the true values and estimates of the model order and noise variance respectively.

## 5.2 ROBOT LOCALIZATION AND MAP BUILDING

Consider a robot that can move on a discrete, two-dimensional grid. Suppose the goal is to learn a map of the environment, which, for simplicity, we can think of as a matrix which stores the color of each grid cell, which can be either black or white. The difficulty is that the color sensors are not perfect (they may accidentally flip bits), nor are the motors (the robot may fail to move in the desired direction with some probability due e.g., to wheel slippage). Consequently, it is easy for the robot to get lost. And when the robot is lost, it does not know what part of the matrix to update. So we are faced with a chicken-and-egg situation: the robot needs to know where it is to learn the map, but needs to know the map to figure out where it is.

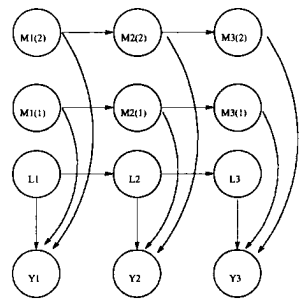

Figure 3: A Factorial HMM with 3 hidden chains. $M_t(i)$ represents the color of grid cell $i$ at time $t$, $L_t$ represents the robot's location, and $Y_t$ the current observation.

The problem of concurrent localization and map learning for mobile robots has been widely studied. In (Murphy 2000), we adopt a Bayesian approach, in which we maintain a belief state over both the location of the robot, $L_t \in \{1, \ldots, N_L\}$, and the color of each grid cell, $M_t(i) \in \{1, \ldots, N_C\}$, $i = 1, \ldots, N_L$, where $N_L$ is the number of cells, and $N_C$ is the number of colors. The DBN we are using is shown in Figure 3. The state space has size $O(N_C^{N_L})$. Note that we can easily handle changing environments, since the map is represented as a random variable, unlike the more common approach, which treats the map as a fixed parameter.

The observation model is $Y_t = f(M_t(L_t))$, where $f(\cdot)$ is a function that flips its binary argument with some fixed probability. In other words, the robot gets to see the color of the cell it is currently at, corrupted by noise: $Y_t$ is a noisy multiplexer with $L_t$ acting as a "gate" node. Note that this conditional independence is not obvious from the graph structure in Figure 3(a), which suggests that all the nodes in each slice should be correlated by virtue of sharing a common observed child, as in a factorial HMM (Ghahramani and Jordan 1997). The extra independence information is encoded in $Y_t$'s distribution, c.f., (Boutilier, Friedman, Goldszmidt and Koller 1996).

The basic idea of the algorithm is to sample $L_{1:t}$ with a PF, and marginalize out the $M_t(i)$ nodes exactly, which can be done efficiently since they are conditionally independent given $L_{1:t}$:

$$P(M_t(1), \ldots, M_t(N_L)|y_{1:t}, L_{1:t}) = \prod_{i=1}^{N_L} P(M_t(i)|y_{1:t}, L_{1:t})$$

Some results on a simple one-dimensional grid world are

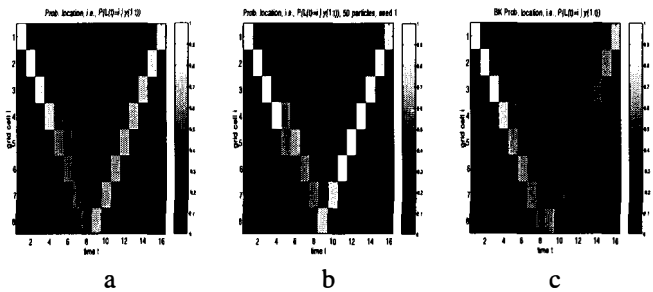

a  b  c

Figure 4: Estimated position as the robot moves from cell 1 to 8 and back. The robot "gets stuck" in cell 4 for two steps in a row on the outgoing leg of the journey (hence the double diagonal), but the robot does not realize this until it reaches the end of the "corridor" at step 9, where it is able to relocalise. (a) Exact inference. (b) RBPF with 50 particles. (c) Fully-factorised BK.

shown in Figure 4. We compared exact Bayesian inference with the RBPF method, and with the fully-factorised version of the Boyen-Koller (BK) algorithm (Boyen and Koller 1998), which represents the belief state as a product of marginals:

$$P(L_t, M_t(1), \ldots, M_t(N_L)|\mathbf{y}_{1:t}) = P(L_t|\mathbf{y}_{1:t}) \prod_{i=1}^{N_L} P(M_t(i)|\mathbf{y}_{1:t})$$

We see that the RBPF results are very similar to the exact results, even with only 50 particles, but that BK gets confused because it ignores correlations between the map cells. We have obtained good results learning a $10 \times 10$ map (so the state space has size $O(2^{100})$) using only 100 particles (the observation model in the 2D case is that the robot observes the colors of all the cells in a $3 \times 3$ neighborhood centered on its current location). For a more detailed discussion of these results, please see (Murphy 2000).

### 5.3 CONCLUSIONS AND EXTENSIONS

RBPFs have been applied to many problems, mostly in the framework of conditionally linear Gaussian state-space models and conditionally finite state-space HMMs. That is, they have been applied to models that, conditionally upon a set of variables (imputed by the PF algorithm), admit a closed-form filtering distribution (Kalman filter in the continuous case and HMM filter in the discrete case). One can also make use of the special structure of the dynamic model under study to perform the calculations efficiently using the junction tree algorithm. For example, if one had evolving trees, one could sample the root nodes with the PF and compute the leaves using the junction tree algorithm. This would result in a substantial computational gain as one only has to sample the root nodes and apply the juction tree to lower dimensional sub-networks.

Although the previoulsy mentioned models are the most famous ones, there exist numerous other dynamic systems admitting finite dimensional filters. That is, the filtering distribution can be estimated in closed-form at any time $t$ using a fixed number of sufficient statistics. These include

- Dynamic models for counting observations (Smith and Miller 1986).

- Dynamic models with a time-varying unknow covariance matrix for the dynamic noise (West and Harrison 1996, Uhlig 1997).

- Classes of the exponential family state space models (Vidoni 1999).

This list is by no means exhaustive. It, however, shows that RBPFs apply to very wide class of dynamic models. Consequently, they have a big role to play in computer vision (where mixtures of Gaussians arise commonly), robotics, speech and dynamic factor analysis.

### References

Akashi, H. and Kumamoto, H. (1977). Random sampling approach to state estimation in switching environments, *Automatica* **13**: 429–434.

Andrieu, C., de Freitas, J. F. G. and Doucet, A. (1999a). Sequential Bayesian estimation and model selection applied to neural networks, *Technical Report CUED/F-INFENG/TR 341*, Cambridge University Engineering Department.

Andrieu, C., de Freitas, J. F. G. and Doucet, A. (1999b). Sequential MCMC for Bayesian model selection, *IEEE Higher Order Statistics Workshop*, Ceasarea, Israel, pp. 130–134.

Becker, A., Bar-Yehuda, R. and Geiger, D. (1999). Random algorithms for the loop cutset problem.

Bernardo, J. M. and Smith, A. F. M. (1994). *Bayesian Theory*, Wiley Series in Applied Probability and Statistics.

Boutilier, C., Friedman, N., Goldszmidt, M. and Koller, D. (1996). Context-specific independence in bayesian networks, *Proc. Conf. Uncertainty in AI*.

Boyen, X. and Koller, D. (1998). Tractable inference for complex stochastic processes, *Proc. Conf. Uncertainty in AI*.

Casella, G. and Robert, C. P. (1996). Rao-Blackwellisation of sampling schemes, *Biometrika* **83**(1): 81–94.

Cowell, R. G., Dawid, A. P., Lauritzen, S. L. and Spiegelhalter, D. J. (1999). *Probabilistic Networks and Expert Systems*, Springer-Verlag, New York.

Crisan, D. and Doucet, A. (2000). Convergence of generalized particle filters, *Technical Report CUED/F-INFENG/TR 381*, Cambridge University Engineering Department.

Crisan, D., Del Moral, P. and Lyons, T. (1999). Discrete filtering using branching and interacting particle systems, *Markov Processes and Related Fields* **5**(3): 293–318.

de Freitas, J. F. G. (1999). *Bayesian Methods for Neural Networks*, PhD thesis, Department of Engineering, Cambridge University, Cambridge, UK.

Dean, T. and Kanazawa, K. (1989). A model for reasoning about persistence and causation, *Artificial Intelligence* **93**(1–2): 1–27.

Doucet, A. (1998). On sequential simulation-based methods for Bayesian filtering, *Technical Report CUED/F-INFENG/TR 310*, Department of Engineering, Cambridge University.

Doucet, A., de Freitas, J. F. G. and Gordon, N. J. (2000). *Sequential Monte Carlo Methods in Practice*, Springer-Verlag.

Doucet, A., Godsill, S. and Andrieu, C. (2000). On sequential Monte Carlo sampling methods for Bayesian filtering, *Statistics and Computing* **10**(3): 197–208.

Doucet, A., Gordon, N. J. and Krishnamurthy, V. (1999). Particle filters for state estimation of jump Markov linear systems, *Technical Report CUED/F-INFENG/TR 359*, Cambridge University Engineering Department.

Ghahramani, Z. and Jordan, M. (1997). Factorial Hidden Markov Models, *Machine Learning* **29**: 245–273.

Gilks, W. R. and Berzuini, C. (1998). Monte Carlo inference for dynamic Bayesian models, Unpublished. Medical Research Council, Cambridge, UK.

Gordon, N. J., Salmond, D. J. and Smith, A. F. M. (1993). Novel approach to nonlinear/non-Gaussian Bayesian state estimation, *IEE Proceedings-F* **140**(2): 107–113.

Green, P. J. (1995). Reversible jump Markov chain Monte Carlo computation and Bayesian model determination, *Biometrika* **82**: 711–732.

Handschin, J. E. and Mayne, D. Q. (1969). Monte Carlo techniques to estimate the conditional expectation in multi-stage non-linear filtering, *International Journal of Control* **9**(5): 547–559.

Isard, M. and Blake, A. (1996). Contour tracking by stochastic propagation of conditional density, *European Conference on Computer Vision*, Cambridge, UK, pp. 343–356.

Kanazawa, K., Koller, D. and Russell, S. (1995). Stochastic simulation algorithms for dynamic probabilistic networks, *Proceedings of the Eleventh Conference on Uncertainty in Artificial Intelligence*, Morgan Kaufmann, pp. 346–351.

Kitagawa, G. (1996). Monte Carlo filter and smoother for non-Gaussian nonlinear state space models, *Journal of Computational and Graphical Statistics* **5**: 1–25.

Kong, A., Liu, J. S. and Wong, W. H. (1994). Sequential imputations and Bayesian missing data problems, *Journal of the American Statistical Association* **89**(425): 278–288.

Liu, J. S. and Chen, R. (1998). Sequential Monte Carlo methods for dynamic systems, *Journal of the American Statistical Association* **93**: 1032–1044.

MacEachern, S. N., Clyde, M. and Liu, J. S. (1999). Sequential importance sampling for nonparametric Bayes models: the next generation, *Canadian Journal of Statistics* **27**: 251–267.

Murphy, K. P. (2000). Bayesian map learning in dynamic environments, *in* S. Solla, T. Leen and K.-R. Müller (eds), *Advances in Neural Information Processing Systems 12*, MIT Press, pp. 1015–1021.

Pearl, J. (1988). *Probabilistic Reasoning in Intelligent Systems: Networks of Plausible Inference*, Morgan Kaufmann.

Pitt, M. K. and Shephard, N. (1999). Filtering via simulation: Auxiliary particle filters, *Journal of the American Statistical Association* **94**(446): 590–599.

Smith, R. L. and Miller, J. E. (1986). Predictive records, *Journal of the Royal Statistical Society* **B 36**: 79–88.

Uhlig, H. (1997). Bayesian vector-autoregressions with stochastic volatility, *Econometrica*.

Vidoni, P. (1999). Exponential family state space models based on a conjugate latent process, *Journal of the Royal Statistical Society* **B 61**: 213–221.

West, M. (1993). Mixture models, Monte Carlo, Bayesian updating and dynamic models, *Computing Science and Statistics* **24**: 325–333.

West, M. and Harrison, J. (1996). *Bayesian Forecasting and Dynamic Linear Models*, Springer-Verlag.